\documentclass[]{fairmeta}
\setcitestyle{numbers,square,comma}

\usepackage[dvipsnames]{xcolor}
\usepackage{xparse}
\usepackage{array}
\usepackage{booktabs}
\usepackage{tabularx}
\usepackage{pifont}
\usepackage{graphicx}
\usepackage{subcaption}

\newcommand{\checkmark}{\textcolor{ForestGreen}{\ding{51}}}
\newcommand{\xmark}{\textcolor{BrickRed}{\ding{55}}}

\newcommand{\dsname}{Embody 3D}
\newcommand{\dshours}{500}
\newcommand{\dspts}{439}

\definecolor{alexcolor}{RGB}{0,102,204}      
\definecolor{andrewcolor}{RGB}{0,153,0}      
\definecolor{stevencolor}{RGB}{204,0,0}      
\definecolor{fabiancolor}{RGB}{255,128,0}    
\definecolor{michaelcolor}{RGB}{153,0,153}   
\definecolor{evonnecolor}{RGB}{255,102,178}  
\definecolor{vasucolor}{RGB}{245,130,48} 

\title{\dsname: A Large-scale Multimodal Motion and Behavior Dataset}

\author{Claire McLean}
\author{Makenzie Meendering}
\author{Tristan Swartz}
\author{Orri Gabbay}
\author{Alexandra Olsen}
\author{Rachel Jacobs}
\author{Nicholas Rosen}
\author{Philippe de Bree}
\author{Tony Garcia}
\author{Gadsden Merrill}
\author{Jake Sandakly}
\author{Julia Buffalini}
\author{Neham Jain}
\author{Steven Krenn}
\author{Moneish Kumar}
\author{Dejan Markovic}
\author{Evonne Ng}
\author{Fabian Prada}
\author{Andrew Saba}
\author{Siwei Zhang}
\author{Vasu Agrawal}
\author{Tim Godisart}
\author{Alexander Richard}
\author{Michael Zollhoefer}

\affiliation{Codec Avatars Lab, Meta}

\contribution{\textit{Author list not sorted by contribution}}

\abstract{
\includegraphics[width=\linewidth]{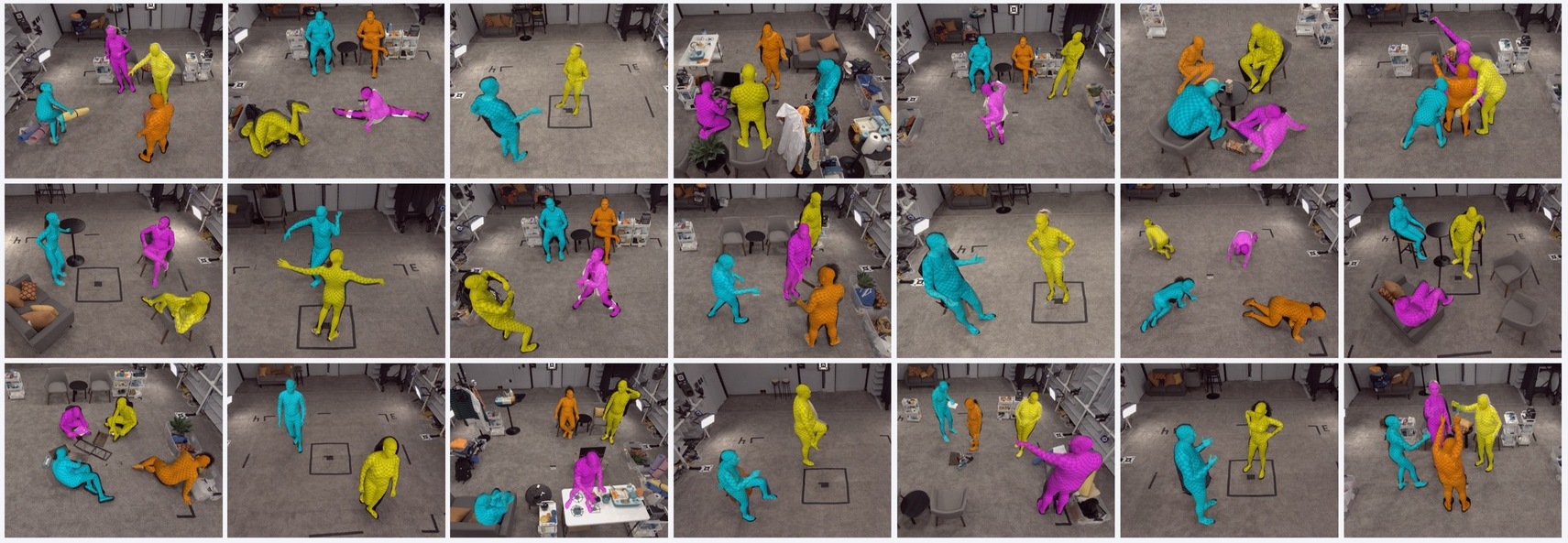}
The Codec Avatars Lab at Meta introduces \dsname, a multimodal dataset of \dshours~individual hours\footnote{\textit{individual hours} refers to number of person-hours in the collection. If we collect one hour of conversation between two participants, it would amount to two individual hours of motion data.} of 3D motion data from \dspts~participants collected in a multi-camera collection stage, amounting to over 54 million frames of tracked 3D motion. The dataset features a wide range of single-person motion data, including prompted motions, hand gestures, and locomotion; as well as multi-person behavioral and conversational data like discussions, conversations in different emotional states, collaborative activities, and co-living scenarios in an apartment-like space. We provide tracked human motion including hand tracking and body shape, text annotations, and a separate audio track for each participant.
}

\date{\today}

\metadata[Dataset]{\url{https://www.meta.com/emerging-tech/codec-avatars/embody-3d}}

\begin{document}

\maketitle

\section{Introduction}
\label{section:intro}

The development of robust motion understanding and synthesis systems critically depends on the availability of high-quality, large-scale motion datasets.
However, current datasets face fundamental trade-offs: they either achieve scale at the expense of quality and completeness~\cite{yi2022generating,lin2023motion,fan2025go,seamless_interaction}, or provide high-quality data in limited quantities~\cite{lu2025humoto,ghorbani2023zeroeggs,IVA:2018,li2021ai,ng2024audio,plappert2016kit,guo2022generating,punnakkal2021babel,lee2019talking,liu2022beat}, see Table~\ref{tab:datasets_overview}.
This limitation has become a significant bottleneck in human motion and behavior research.

\textbf{Limitations of 2D motion datasets} lie in their quality and comprehensiveness.
While it is easy to scale 2D video data to thousands of hours, it is challenging to convert such data into high quality 3D motion data.
Monocular tracking suffers from depth ambiguity, motion blur, limited resolution around critical areas like the hands, occlusions and partial visibility of the human in the frames, as well as the inability to establish a common 3D world space when multiple actors appear in a video.
While 2D datasets can provide breadth, they lack the precision and spatial consistency required for many applications.

\textbf{3D motion datasets} address the quality concerns through sophisticated multi-view systems that enable accurate 3D tracking.
However, scaling collections in such specialized, complex systems is costly, such that existing 3D datasets remain relatively small in volume.
Furthermore, even among 3D datasets, completeness is often compromised: many existing datasets fail to provide hand tracking or body shapes.

\textbf{Domain specialization} plagues both 2D and 3D datasets.
Even large scale video-based datasets typically focus on either conversations~\cite{seamless_interaction} or locomotion~\cite{fan2025go}, but rarely both.
In smaller scale 3D motion datasets, this task specialization is even more severe, rendering them unfit to model generic human motion and behavior.
Additionally, existing datasets frequently lack critical modalities that are strongly linked to human motion, such as speech or text annotations.

\textbf{\dsname} aims to overcome these limitations and provides a single, large-scale dataset with comprehensive coverage of human motion and behavior.
Our dataset consists of 500 individual hours of human motion, tracked in a multi-view collection system with 80 cameras with 24 mega-pixels each.
We provide full body tracking including hands and body shape, together with additional data modalities such as audio and text annotations.
Instead of specializing on a single task, we provide motion and behavior data on a comprehensive set of tasks, including single-person collections such as charades, locomotion, or hand interactions, as well as multi-person collections like conversations, collaborative activities, furniture and object interactions, and co-living scenarios.
Overall, the dataset has more than 54 million 3D motion frames and \dspts~participants.
We provide body and hand tracking as well as body shape parameters in SMPL-X \cite{SMPL-X:2019} format, audio that is separated per participant using beamforming over a 640-channel microphone array, fine-grained text annotations created by human annotators, and segment-level annotations through prompting (like \textit{play tennis} for charades or \textit{be angry} for conversational settings).

\begin{table}[tbh]
    \newcolumntype{Y}{>{\centering\arraybackslash}m{0.054\textwidth}} 
    \centering
    \footnotesize
    \begin{tabularx}{1.0\textwidth}{XYYYYYYYYYY}
        \toprule
        & \multicolumn{2}{c}{\textbf{dataset size}}
        & \multicolumn{3}{c}{\textbf{tracking}}
        & \multicolumn{2}{c}{\textbf{data type}}
        & \multicolumn{2}{c}{\textbf{modalities}} \\
        \cmidrule(lr){2-3}
        \cmidrule(lr){4-6}
        \cmidrule(lr){7-8}
        \cmidrule(lr){9-10}
        & \shortstack{\textbf{individual}\\\textbf{hours}}
        & \textbf{subjects}
        & \shortstack{\textbf{full}\\\textbf{body}}
        & \textbf{shape}
        & \textbf{hands}
        & \shortstack{\textbf{loco-}\\\textbf{motion}}
        & \shortstack{\textbf{conver-}\\\textbf{sation}}
        & \textbf{audio}
        & \textbf{text}
        & \shortstack{\textbf{multi-}\\\textbf{person}} \\
        \midrule
        \midrule
        \multicolumn{11}{c}{\textbf{2D motion datasets (monocular tracking from video)}} \\
        \midrule
        SHOW~\cite{yi2022generating} & 27 & 4 & \xmark & \checkmark & \checkmark & \xmark & \checkmark & \checkmark & \xmark & \xmark \\
        Motion-X~\cite{lin2023motion} & 144 & - & \checkmark & \checkmark & \checkmark & \checkmark & \xmark & \xmark & \checkmark & \xmark \\
        MotionMillion~\cite{fan2025go} & 2,000 & - & \checkmark & \checkmark & \xmark & \checkmark & \xmark & \xmark & \checkmark & \checkmark \\
        Seamless Interaction~\cite{seamless_interaction} & 8,130 & 4,284 & \xmark & \xmark & \checkmark & \xmark & \checkmark & \checkmark & \checkmark & \checkmark \\
        \midrule
        \midrule
        \multicolumn{11}{c}{\textbf{3D motion datasets}} \\
        \midrule
        HUMOTO~\cite{lu2025humoto} & 2 & 1  & \checkmark & \xmark & \checkmark & \checkmark & \xmark & \xmark & \checkmark & \xmark \\
        ZeroEGGS~\cite{ghorbani2023zeroeggs} & 2 & 1  & \checkmark & \xmark & \checkmark & \xmark & \checkmark & \checkmark & \xmark & \xmark \\
        Trinity~\cite{IVA:2018} & 4 & 1  & \checkmark & \xmark & \xmark & \xmark & \checkmark & \checkmark & \xmark & \xmark \\
        AIST++~\cite{li2021ai} & 5 & 30  & \checkmark & \xmark & \xmark & \checkmark & \xmark & \checkmark & \xmark & \xmark \\
        Audio2Photoreal~\cite{ng2024audio} & 8 & 4 & \checkmark & \checkmark & \checkmark & \xmark & \checkmark & \checkmark & \xmark & \checkmark \\
        KIT-ML~\cite{plappert2016kit} & 11 & 111 & \checkmark & \xmark & \xmark & \checkmark & \xmark & \xmark & \checkmark & \xmark \\
        HumanML3D~\cite{guo2022generating} & 29 & - & \checkmark & \checkmark & \xmark & \checkmark & \xmark & \xmark & \checkmark & \xmark \\
        BABEL~\cite{punnakkal2021babel} & 43 & 346  & \checkmark & \xmark & \xmark & \checkmark & \xmark & \xmark & \checkmark & \xmark \\
        BEAT~\cite{liu2022beat} & 76 & 30 & \checkmark & \checkmark & \checkmark & \xmark & \checkmark & \checkmark & \xmark & \xmark \\
        Talking with Hands~\cite{lee2019talking} & 100 & 50 & \checkmark & \xmark & \checkmark & \xmark & \checkmark & \checkmark & \xmark & \checkmark \\
        \midrule
        \textbf{\dsname} & \dshours & \dspts & \checkmark & \checkmark & \checkmark & \checkmark & \checkmark & \checkmark & \checkmark & \checkmark \\
        \bottomrule
    \end{tabularx}
    \caption{\textbf{Overview of existing datasets.}
    2D motion datasets scale in volume, but suffer from low tracking quality, depth ambiguity, and lack of a consistent 3D space.
    3D motion datasets are of high tracking quality and grounded in a defined 3D world space, but typically are limited in size.
    All existing datasets lack important aspects of human motion (hands, shape, upper and lower body tracking), types of motion (locomotion vs.\ conversational, gesture-focused motion), interactions between multiple people, or additional modalities like audio or text annotations.
    \dsname~provides the first comprehensive high quality 3D motion dataset that checks all these boxes.}
    \label{tab:datasets_overview}
\end{table}

\section{Dataset Details}
\label{section:dataset}

Our dataset features seven different subcategories of motion with different properties.

\textbf{Charades} contains prompted motion.
Participants are instructed to perform a specific motion, like jumping or shooting an arrow, and are being recorded for 15 seconds per motion prompt.
We provide the tracked motion and the text prompt for the motion.
Overall, this subcategory has 88.9h of motion data from a total of 221 participants.

\textbf{Hand Interactions.} A portion of the dataset is collected with special emphasis on hand motion and hand-body interactions.
Each participant was instructed to perform several hand and arm motions. Many of these motions have self-contact between both hands or between the hands and body.
In total, the dataset has 111.3h of hand-focused motion segments from a total of 137 participants.

\textbf{Locomotion}. In this category, participants were instructed to move in a specific way, such as different styles of jumping, walking, or running.
The dataset has a total of 21h of locomotion-focused data from 46 participants.

\textbf{Dyadic Conversations.} Humans are social animals, and as such communication between humans plays a large role in our everyday lives.
To enable building authentic conversational virtual humans, we put a special emphasis on conversational collections.
We collected a total of 59.4 individual hours of such conversational motion data with a total of 86 participants.
The conversations were guided by instructions and prompts.
Participants were asked to have conversations about various topics in different emotions like anger, happiness, sadness and others, as well as unguided free-form conversations.

\textbf{Multi-Person Conversations.} Moving beyond single person and dyadic data, the dataset offers 125.2 individual hours of multi-person conversations with a total of 210 participants.
Besides extending the dyadic case to multiple participants, we also add \textbf{furniture interaction} to the data collection and allow participants to use chairs, a high-table, or a couch as opportunities to sit down during their conversations.

\textbf{Scenarios} is a subcategory in which multiple participants perform a given scenario or a collaborative or competitive activity.
These scenarios include playing games, assembling furniture, or competing on given tasks. 
Notably, this section contains object and furniture interactions.
Overall, we provide 49.2 individual hours of motion data from 77 participants in this subcategory.

\textbf{Day in the Life} is the most challenging motion subcategory in our dataset.
Three to four participants interact with each other in a small apartment-like setup with different objects and furniture.
We instructed participants with different goals that focus around typical co-living, hosting, or group activity scenarios.
In total, this category consists of 46.4 individual hours of motion data from a total of 77 participants.

\begin{figure}[htbp]
    \captionsetup[sub]{skip=2pt}
    \captionsetup{skip=4pt}
    \centering
    \begin{subfigure}[t]{0.13\textwidth}
        \centering
        \includegraphics[width=\linewidth]{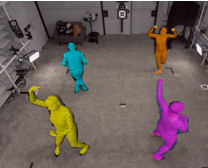}
        \caption*{\parbox{\linewidth}{\centering\footnotesize Charades}}
    \end{subfigure}
    \hfill
    \begin{subfigure}[t]{0.13\textwidth}
        \centering
        \includegraphics[width=\linewidth]{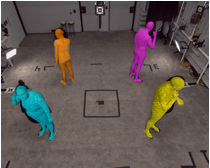}
        \caption*{\parbox{\linewidth}{\centering\footnotesize Hand Interactions}}
    \end{subfigure}
    \hfill
    \begin{subfigure}[t]{0.13\textwidth}
        \centering
        \includegraphics[width=\linewidth]{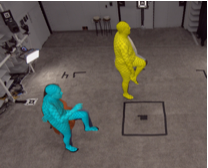}
        \caption*{\parbox{\linewidth}{\centering\footnotesize Locomotion}}
    \end{subfigure}
    \hfill
    \begin{subfigure}[t]{0.13\textwidth}
        \centering
        \includegraphics[width=\linewidth]{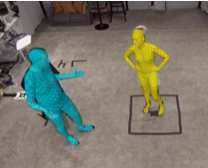}
        \caption*{\parbox{\linewidth}{\centering\footnotesize Dyadic Conversations}}
    \end{subfigure}
    \hfill
    \begin{subfigure}[t]{0.13\textwidth}
        \centering
        \includegraphics[width=\linewidth]{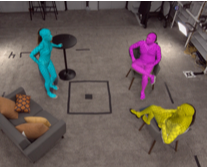}
        \caption*{\parbox{\linewidth}{\centering\footnotesize Multi-person Conversations}}
    \end{subfigure}
    \hfill
    \begin{subfigure}[t]{0.13\textwidth}
        \centering
        \includegraphics[width=\linewidth]{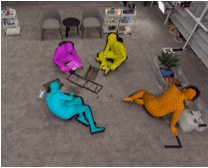}
        \caption*{\parbox{\linewidth}{\centering\footnotesize Scenarios}}
    \end{subfigure}
    \hfill
    \begin{subfigure}[t]{0.13\textwidth}
        \centering
        \includegraphics[width=\linewidth]{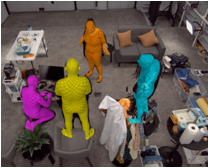}
        \caption*{\parbox{\linewidth}{\centering\footnotesize Day in the Life}}
    \end{subfigure}
    \caption{
        Example frames from each dataset subcategory.
        For single-person categories like Charades, Hand Interactions, and Locomotion, we collect multiple participants at the same time for better throughput in terms of individual participant hours. For interactive subcategories like conversations or scenarios, on the contrary, all participants in the scene interact with each other.
    }
    \label{fig:seven_panels}
\end{figure}

Table~\ref{tab:subcategories} provides a summary of amount of data and available assets per subcategory.
Figure~\ref{fig:seven_panels} shows an example frame from each of the subcategories.

\begin{table}[tbh]
    \newcolumntype{Y}{>{\centering\arraybackslash}m{0.085\textwidth}} 
    \centering
    \footnotesize
    \begin{tabularx}{0.7\textwidth}{XYYYY}
        \toprule
        & \shortstack{\textbf{hours of}\\\textbf{motion data}} 
        & \shortstack{\textbf{number of}\\\textbf{participants}} 
        & \textbf{audio} 
        & \shortstack{\textbf{text}\\\textbf{annotations}} \\
        \midrule
        Charades                   &  88.9   & 221   & \xmark     & \checkmark \\
        Hand Interactions          & 111.3   & 137   & \xmark     & \xmark \\
        Locomotion                 &  21.0   &  46   & \xmark     & (\checkmark) \\
        Dyadic Conversations       &  59.4   &  86   & \checkmark & (\checkmark) \\
        Multi-person Conversations & 125.2   & 210   & \checkmark & \xmark \\
        Scenarios                  &  49.2   &  77   & \checkmark & \checkmark \\
        Day in the Life            &  46.4   &  77   & \checkmark & \checkmark \\
        \bottomrule
    \end{tabularx}
    \caption{Available data per subcategory of the dataset. (\checkmark) refers to high-level information only, for instance a dyadic conversations might have an emotion label but no fine-grained text annotations.}
    \label{tab:subcategories}
\end{table}

\section{Collection System and Dataset Acquisition}
\label{section:corpus}

\textbf{System.} The \dsname~collection system is a multimodal collection system of $6$m by $6$m by $3.6$m high.
The capture area covered by cameras is $3.6$m by $3.6$m.
The room is outfitted with custom anechoic acoustic treatment and a pipe grid system to mount equipment.
The camera array uses high-end global shutter machine vision cameras with a 24.47-megapixel resolution (5320 x 4600) as collect data at 30fps.
There are 80 cameras distributed around the covered volume. The 64 body-tracking cameras are equipped with 8–15mm F4 EF lenses, and the exposure is set to 4 milliseconds.
The 16 face-tracking cameras are equipped with 35mm F1.4 EF lenses, and the exposure is set to 2 milliseconds.
We arranged 14 LED panels, equally distributed throughout the collection system, achieving an average illuminance of approximately 650 lux throughout the volume, similar to a bright indoor room.
The microphone system is made from 5 custom, in-house-designed MEMS microphone arrays.
Each array consists of 128 MEMS microphone elements arranged in a spherical array, effectively recording 10th-order ambisonics.
The five 128-mic MEMS arrays combine to create a larger array, totaling 640 audio channels.

\textbf{Data collection.} All data collection sessions in this dataset have been supervised by research assistants who ensure participants are properly prepared and briefed for the tasks they are given during the collection.
Prior to the collection, each participant has been informed about the use of their data for research purposes and has signed a consent form.
All participants complete a calibration session for body shape estimation.
Research assistants then guide the participants through their sessions, give instructions, and provide feedback.
Additionally, each recording is monitored in real time by a research assistant to flag quality issues.

\textbf{Text Annotations.} We provide detailed, human-generated text annotations for all segments from the Scenarios and Day-in-the-Life subcategories.
These annotations include scene level information that describe the scene on a higher level, as well as detailed pose and motion annotations for each person in the scene.
We additionally asked annotators to assign labels to each participant that describe their emotional state based on their facial expression, their pose, and their speech.

\section{Data Processing}
\label{section:corpus:processing}

\begin{figure}
    \captionsetup{skip=4pt}
    \centering
    \includegraphics[width=0.9\linewidth]{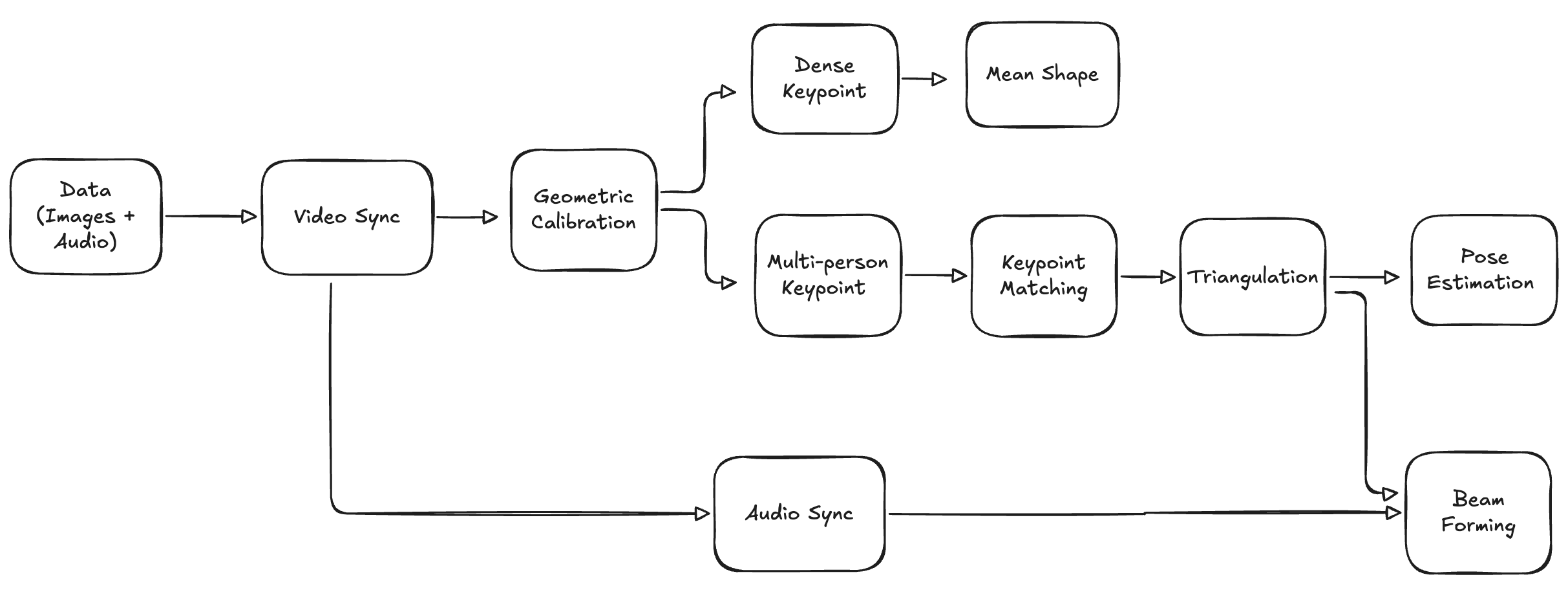}
    \caption{End to end processing pipeline for generating multi-person poses and audio from raw collected data.}
    \label{fig:pipeline}
\end{figure}

In the following, we describe the data processing pipeline from raw collected data to tracked 3D motion, body shapes, and speech-separated audio channels.
A schematic overview of the process is illustrated in Figure~\ref{fig:pipeline}.

\subsection{Synchronization and Calibration}

\textbf{Multi-Camera and Audio Synchronization.}
During a recording session, timestamps for each camera frame are stored alongside the raw image data.
While all cameras are co-triggered by the same clock and trigger source, frames still need to be synchronized in post to handle dropped frames.
The timestamps are combined across all the cameras and clustered to generate a global frame list.
Any cameras whose timestamps are above a certain threshold from the median timestamp for each cluster are filtered out. Additionally, any frames where too few cameras are present are also filtered out. This leaves a global frame mapping and a final list of cameras that are in sync with one another. An encoding of the timestamp is embedded into the audio files, allowing us to synchronize the audio stream to the camera frames.
Each collection is recorded in a series of segments.
For each segment, we take the start and end timestamps generated by the camera synchronization stage and synchronize to the audio samples.
Frame drops in the audio stream are replaced with zeros.

\textbf{Multi-Camera Geometric Calibration.}
For multi-camera calibration, we developed a custom fiducial tracking board mounted on a wheeled cart.
The operator moves the fiducial tracking board around the room to build an extrinsic connection graph.
We also built a smaller fiducial tracking rig for camera intrinsic calibration.
On average, our p50 re-projection error is under 0.2px, with p99 error about 0.8px on average.
We mounted a custom fiducial tracking system onto the microphone array systems to align the microphone coordinate system with the camera coordinate system.
Finally, floor fiducial targets were added to estimate the floor plane.
Microphone positions and floor tags are detected and triangulated using RANSAC and PnP pose estimation.
Finally, a plane is fit to the detected floor points, providing a floor-centric world coordinate system that is consistent among all collection sessions.

\subsection{Participant Shape Estimation}

\begin{figure}[htbp]
    \captionsetup[sub]{skip=2pt}
    \captionsetup{skip=4pt}
    \centering
    \begin{subfigure}[t]{0.25\textwidth}
        \centering
        \includegraphics[width=\linewidth]{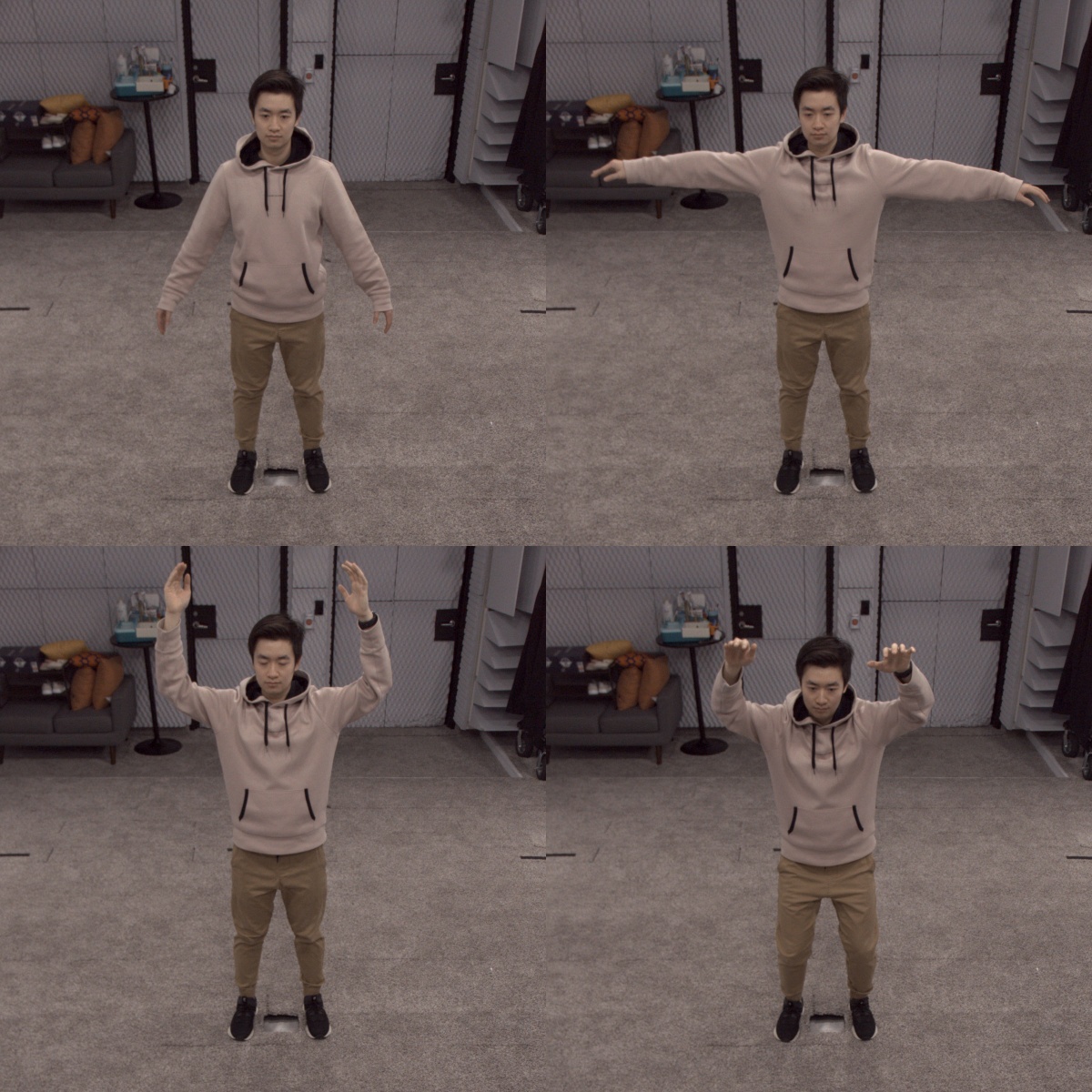}
        \caption*{\parbox{\linewidth}{\centering\footnotesize Calibration Poses}}
    \end{subfigure}
    \begin{subfigure}[t]{0.15\textwidth}
        \centering
        \includegraphics[width=\linewidth]{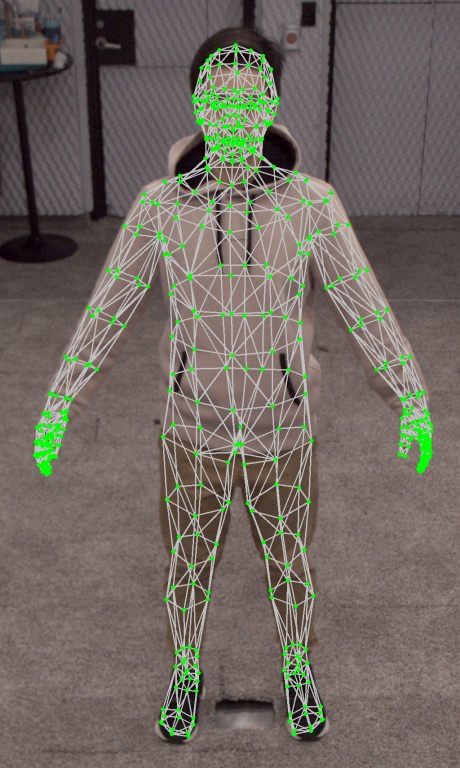}
        \caption*{\parbox{\linewidth}{\centering\footnotesize Dense Keypoint Detection}}
    \end{subfigure}
    \begin{subfigure}[t]{0.15\textwidth}
        \centering
        \includegraphics[width=\linewidth]{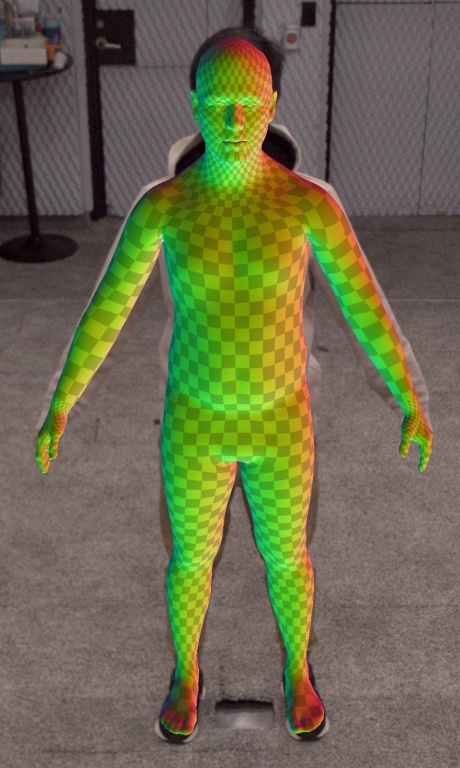}
        \caption*{\parbox{\linewidth}{\centering\footnotesize Body Shape Estimation}}
    \end{subfigure}
    \caption{Each participant performs calibration poses to obtain person specific body shape and reference face images.  
    }
    \label{fig:participant_calibration}
\end{figure}

During the collections, we ask participants to perform a series of four calibration poses, from which we extract their mean body shape. These calibration poses comprise A, T, C, and T-Rex poses. Each of these poses are performed by a single participant at a time for a few seconds per pose.
On these calibration poses, we run a dense keypoint model and optimize shape coefficients of the linear 3D human shape model. See Figure~\ref{fig:participant_calibration} for a reference example.

\subsection{Multi-Person Pose Estimation}

\begin{figure}[htbp]
    \centering

    \begin{subfigure}[t]{0.45\textwidth}
        \centering
        \includegraphics[width=\linewidth]{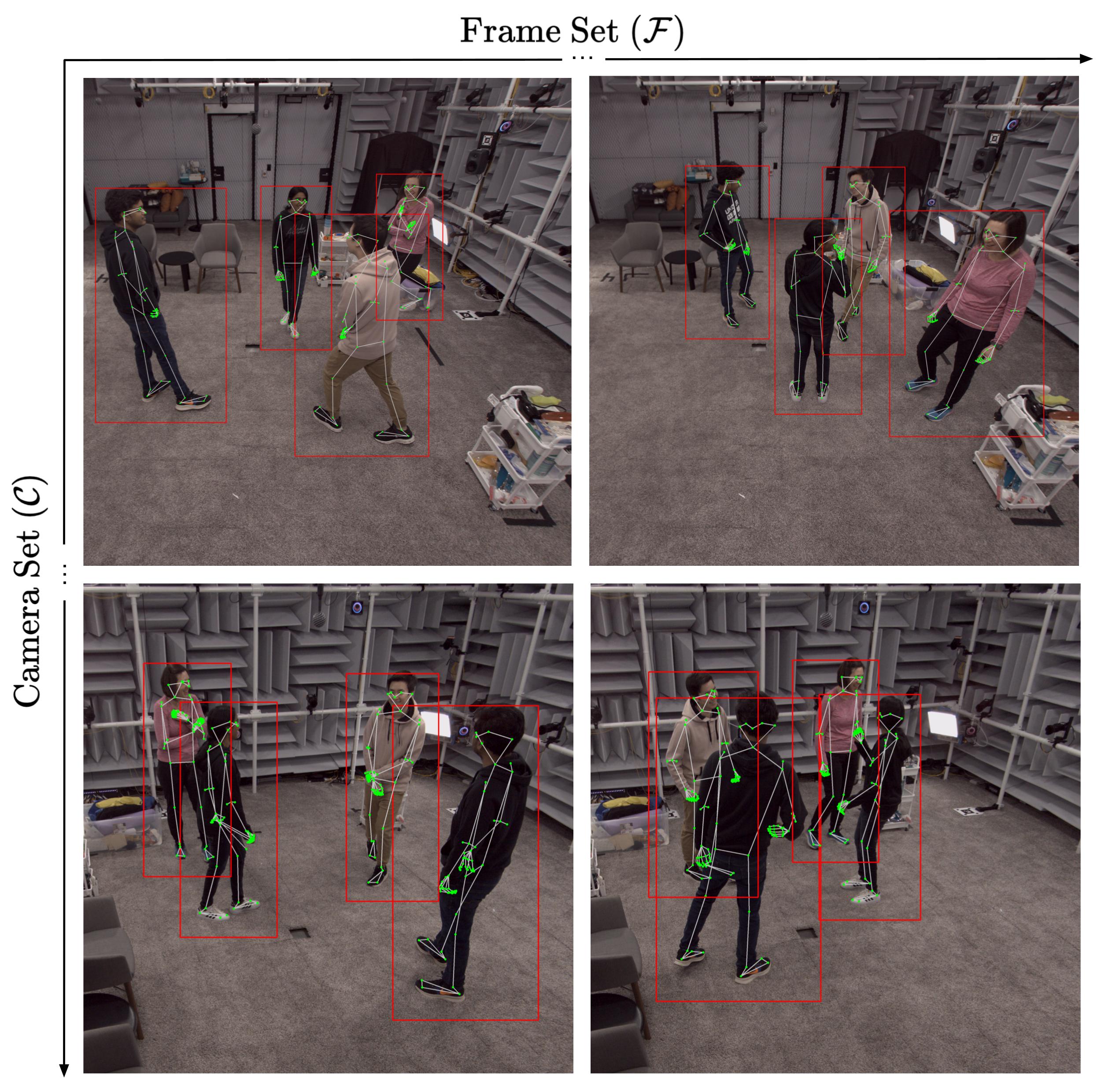}
        \caption{Keypoint Detection}
    \end{subfigure}
    \begin{subfigure}[t]{0.45\textwidth}
        \centering
        \includegraphics[width=\linewidth]{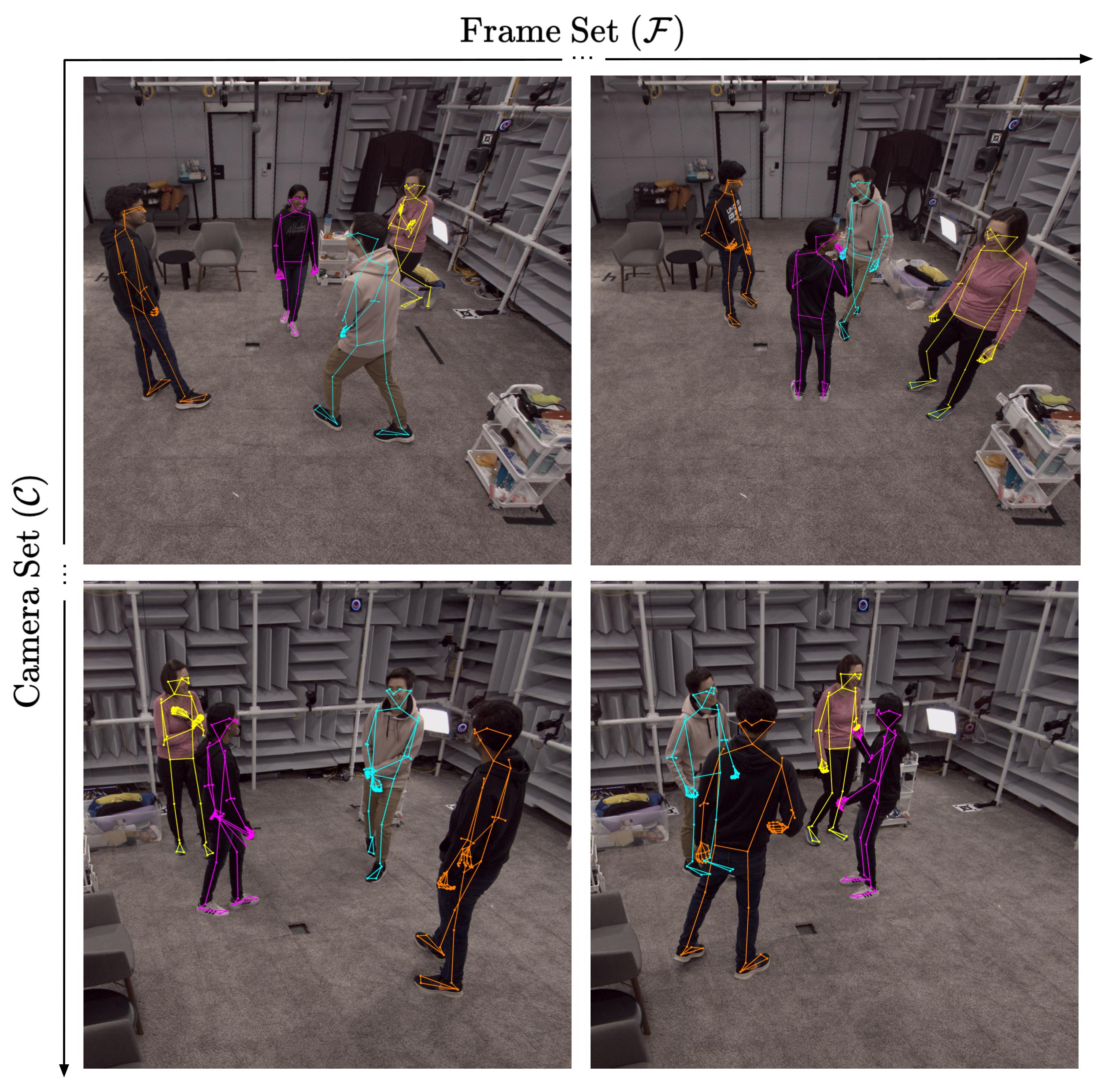}
        \caption{Keypoint Matching}
    \end{subfigure}
    \hfill
    \caption{ We generate 2d keypoint detections within the bounding boxes of  a multi-person detector (left). Keypoints are then matched to participants using geometric and appearance cues (right). 
    }
    \label{fig:multi_person_keypoints_and_matching}
    \vspace{-0.5cm}
\end{figure}

\textbf{Multi-Person Keypoint Detection.}
In each image from each synchronized camera, we run a person bounding box detector. We sort by detection confidence and keep the top $N$ bounding boxes, where $N$ is the number of participants for a given collection session. For each bounding box instance, we run the Sapiens-1B keypoint pose detector model~\cite{sapiens}, providing 308 keypoints for face and body. Example outputs can be seen in Figure~\ref{fig:multi_person_keypoints_and_matching}a.

\textbf{Keypoint Matching.}
Given the set of multi-person keypoints, the matching problem consists of finding the detections associated with each participant.
We solve the matching problem following a bottom-up approach.
First we create per-frame spatial clusters (one for each participant) by lifting 2D detections into ray bundles and grouping based on ray-to-ray distance.
Then, we propagate clusters across consecutive frames by comparing similarity of their respective 2D detections.
This give us spatio-temporal clusters of 2D keypoints.
We match these spatio-temporal clusters with participants using an off-the-shelf face embedding model~\cite{ibug}.
Within each cluster, we sample five face crops from images with the highest detection confidence.
Then, we compute cosine similarity between the embeddings of a reference face image from each participant and the embeddings of cluster samples.
We match identities by applying the Hungarian algorithm to the similarity matrix of face embeddings.
See Figure~\ref{fig:multi_person_keypoints_and_matching}b for an example of keypoint matching.

\textbf{Keypoint Triangulation.}
We obtain initial estimations of 3D keypoints from RANSAC: for each keypoint we compute triangulations from random camera pairs and keep the one with lowest reprojection error on the remaining cameras.
We further refine the keypoint positions by minimizing an energy that incorporates point-to-ray distance, temporal smoothness, and bone length constrains.

\textbf{Pose Tracking.}
To obtain the skeleton state at each frame we train a pose encoder model mapping shape and 3D keypoints to joint rotations.
During training we minimize the distance between 3D keypoints and joint positions~\cite{SMPL:2015}.
Additionally, we regularize joint rotations through a pre-trained pose prior model.
The pose encoder model consists of a Procrustes module that aligns torso joints to 3D keypoints, followed by refinement modules mapping joint-to-keypoint 3D residuals into joint rotation residuals~\cite{Carreira2015HumanPE}. 

\subsection{Beamforming}

We developed a beamforming algorithm over the 640 MEMS microphone channels to separate speech from each participant in the scene.
Based on human annotations for noise, bleed, and distortions of the beamforming algorithm, we optimized its hyper-parameters to achieve best separation while minimizing distortion.
We release the non-separated audio from a reference microphone in the center of the collection system as well as a separated speech channel for each participant in each segment.

\subsection{Quality Assurance}

Human annotators reviewed the entire dataset to ensure high tracking quality.
Each participant's tracked motion has been overlaid with the original video data from four different camera views to spot tracking errors, severe jitter, or other inconsistencies.
Annotators scored tracking quality and accuracy on a Likert scale from one to five and we discarded all segments with an average score of lower than $2.5$.
We observed annotator ratings above $2.5$ to be free of significant errors, and rating penalties mostly being rooted in minor misalignments in subtle details.

\clearpage
\newpage
\bibliographystyle{assets/plainnat}
\bibliography{paper}

\end{document}